# Pupil Center Detection Approaches: A comparative analysis

Talía Vázquez Romaguera[1], Liset Vázquez Romaguera[2], David Castro Piñol [3],
Carlos Román Vázquez Seisdedos[1]

[1]*Universidad de Oriente*, Center for Neuroscience Studies, Images and Signals Processing,
Santiago de Cuba, Cuba

[2]École Polytechnique de Montréal, Electrical Engineering Department, Montreal, Canada

[3] *Universidad de Oriente*, Department of Telecommunications, Santiago de Cuba, Cuba

{tvazquez, cvazquez}@uo.edu.cu, liset.vazquez@polymtl.ca, davidpinol@uo.edu.cu

**Abstract.** In the last decade, the development of technologies and tools for eye tracking has been a constantly growing area. Detecting the center of the pupil, using image processing techniques, has been an essential step in this process. A large number of techniques have been proposed for pupil center detection using both traditional image processing and machine learning-based methods. Despite the large number of methods proposed, no comparative work on their performance was found, using the same images and performance metrics. In this work, we aim at comparing four of the most frequently cited traditional methods for pupil center detection in terms of accuracy, robustness, and computational cost. These methods are based on the circular Hough transform, ellipse fitting, Daugman's integro-differential operator and radial symmetry transform. The comparative analysis was performed with 800 infrared images from the CASIA-IrisV3 and CASIA-IrisV4 databases containing various types of disturbances. The best performance was obtained by the method based on the radial symmetry transform with an accuracy and average robustness higher than 94%. The shortest processing time, obtained with the ellipse fitting method, was 0.06 s.

## 1 Introduction

Eye-tracking or gaze position systems are widely used in a large number of applications: medical research (psychological, neurophysiological, cognitive, ophthalmological) (1–3), rehabilitation (4), driver evaluation and fatigue detection (5), marketing and usability studies (6), help for the disabled people (7), video games (8) and human-computer interaction (9), among others. Currently, the systems used in humans are based on the following non-invasive techniques: electro-oculography which uses pairs of electrodes placed around the eye, and video-oculography which uses video cameras to capture images of the eyes. The main drawbacks of electro-oculography are its low immunity to disturbances (interferences, drifts, and noise) in the acquisition system due to the low signal level (50 to 3500 µV) and the high cost required to guarantee the electrical safety of the system. On the other hand, the video-oculography (VOG) has greater mobility and adaptability as it uses small, lightweight, reusable, safe, and relatively low-cost devices. Therefore, in recent years, its use has increased. The main challenge of this technique focuses on the processing methods that include the following steps:(a) face location, (b) eye location,(c) center pupil detection and (d) calculation of the gaze direction. The first step depends on the quality and assembly of the cameras used, which are usually infrared (IR). This imaging modality offers better contrast between the iris and the pupil. The most widely used VOG systems for medical applications, which is the interest of this work, are those that use chin rests with a coupled camera that focuses on one eye (monocular) or both (binoculars). The step (a) is not required in this configuration. Once the subject is positioned on the chin rest, and the camera is focused on his eye, step (b) is relatively easy to implement and does not require a high precision, because only the eye should be framed, so that the pupil is clearly distinguished. Step (c) is the most complex and important because of its accuracy and precision conditions step (d) and the subsequent application made by the software of these systems. Therefore, it is the great challenge of these systems and is the main focus of this paper.

In recent years, a large number of pupil center detection procedures have been proposed using both traditional image processing and machine learning-based methods. The latter, also called appearance-based methods (10), estimate the pupil center from features of its appearance when the subject looks at a specific point in the scene. It requires prior training and therefore a high number of images, computational resources and time. Furthermore, traditional methods are subdivided into two groups: those based on characteristics and those based on models. Characteristic-based methods estimate the pupil center using various image processing functions, which segment the edge of the pupil according to its characteristics (resolution, contrast, color, etc.), and then estimate the center of gravity. Another approach is to analyze, through a mathematical formulation, the relationship between the orientations of the image gradient vectors and the position where they intersect most frequently (possible center of the pupil). Model-based methods estimate the pupil as the center of the geometric model that best matches the shape of the edge of the pupil. Depending on the angular position of the eye, the model can be circular or elliptical. These methods do not need training and for this reason, they are the ones addressed in this work.

Among all the traditional methods, the most cited are those based on the Circular Hough Transform (11), the ellipse fitting (12), Daugman's integro-differential operator (13) and the Radial Symmetry Transform (14). In a search carried out in Google Scholar, they have 8620, 8500, 4175, and 750 citations, respectively.

In (15) a method was proposed that uses the Circular Hough Transform in a dataset of 52 IR images of the same subject with the gaze oriented in all directions. The accuracy metric used was the percent relative error (ratio between Euclidean distance and pupil diameter) and no metric was reported for computational cost. Likewise, in (16) a similar method was evaluated with 1000 IR images of 12 subjects, with variations in lighting, reflections, eyelash interference, and blurring. Although the accuracy was expressed in %, its definition does not appear. Similarly, the processing time was reported without further explanation on how it was obtained.

In (17) a method based on the ellipse fitting was proposed after performing the decomposition of contours into sinusoidal components of the binary image. The evaluation was performed with 53926 IR images acquired from the right eye during a walk with variable lighting conditions. In the dataset, 74 images with low visibility of the pupil due to flickering, saccades, and distortion caused by excessive lighting were discarded. Also, 500 randomly selected IR images from the "CASIA-Iris-Thousand" database (18) were used. The accuracy metric was the modular Euclidean distance, expressed in %, and the computational cost was the detection time, per image, in ms. In (19) an ellipse fitting method was applied, after the detection of edges using the Canny algorithm (20). The evaluation was performed with 130,856 IR images from the "Labeled Pupils in the Wild" database (21). This dataset contains images from subjects of several ethnicities, under variable lighting conditions, use of eyeglasses, contact lenses, makeup and with different gaze directions. The accuracy metric was the ratio defined as the percentage detection of correctly detected pupils, when the Euclidean distance is less than or equal to 5 pixels. The computational cost was expressed in terms of the detection time per image, in ms.

In 2004, Daugman proposed the integro-differential operator-based method for detecting the contour of the pupil and the iris (22), which assumes that the edges of the pupil and the iris are circular in shape. It seeks, in a smoothed image by a Gaussian filter, the center and radius of a bordered circle on which, the integral derivative is maximum. In (23) an optimized version was evaluated with 756 IR images from the CASIA V1 database, all with good contrast between the pupil, iris, and sclera regions. In (24) the integro-differential operator was combined with the Hough transform. It was evaluated using

756 images from CASIA V1 and CASIA V2 databases (the number was not specified). In both works, the results were presented qualitatively and no metrics were reported for accuracy and computational cost.

In (25) a method based on the radial symmetry transform was proposed to identify regions of interest with radial symmetry within a scene. In (26) this method was applied for pupil detection and was evaluated using 1295 IR images captured from six volunteers. Of these, 410 are sharp and in the rest there are presences of eyelashes, eyelids, eyeglasses, and bright spots. The accuracy metric (in %) was obtained by subtracting, from 100, the value of the relative percentage error (ratio between the Euclidean distance and the radius of the pupil). The computational cost was measured as the detection time per image, in ms.

From the previous review, it can be seen that, in the different proposals, there is no uniformity between the image databases, the computational resources, and the evaluation metrics used. After extensive searching (Google Scholar, IEEE Explorer, ScienceDirect, Springer Link, and ACM Digital Library), no comparative work was found on the performance of methods for pupil center detection. Some works compared their proposed method with another existing one(s) [26],[27] using images that were not acquired or processed under equal conditions and whose performance metrics were different.

The goal of this paper is to perform a comparative analysis of four of the most cited traditional methods for detecting the pupil center, using the same images and performance metrics for all the methods.

This paper is organized as follows. The "Materials and Methods" section describes the images and algorithms that support the research. In the "Results and Discussion" section, the evaluation methodology is explained and the results obtained are analyzed comparatively. Finally, the conclusions of the work are exposed.

## 2 Materials and methods

The pupil location procedure is a complex task, since the shape of the object to be segmented is not necessarily regular or its limits are not always well defined. Generally, an area of interest is obtained first to facilitate pupil location and so that the algorithms work faster. In this work, it is assumed that this step was previously performed, so that the focus is on locating the center of the pupil. Next, the four methods to be compared are described.

### 2.1 Circular Hough Transform

Hough transform was proposed by Paul Hough to find curves (lines, polynomials, circles, and others) in digitized images. It is based on the projection of an *N*-dimensional image space to another space of parameters of dimension *M* (Hough space), which are related through a mathematical model. The transform is mainly used in two and three dimensions, to find lines, parabolas, centers of circles with fixed radius and variables, since for larger dimensions, the number of variables, the complexity of the algorithm and the computational cost increase considerably. The Circular Hough Transform (CHT) is a particular case when the mathematical model between both spaces is represented by the function *g* of a circle expressed as:

$$g(x_j, y_j, x_0, y_0, r) = (x_j - x_0)^2 + (y_j - y_0)^2 - r^2 \quad (1)$$

where $(x_0, y_0)$ are the coordinates of the center of the circle, of radius *r*. In this case, the parameter space is three-dimensional, that is, it has 3 parameters: two for the center of the circle and one for the radius.

According to the CHT (expression 1), each pixel in the image space corresponds to a circle in the Hough space and vice versa. All points of the edge of a circle in the image space are transformed into several circles with the same radius *r* (Fig. 1.a). The intersection of these circles determines the center of the circle detected in Hough space whose coordinates are $(x_0, y_0)$.

The CHT result is stored in an image-size matrix called "Accumulator Array" (Figure 1.b). The accumulator value is updated (increasing by 1) for each circle generated by using the CHT. The maximum accumulator value represents the center O of the detected circle and is obtained when all the circles generated by the edge pixels vote (intersect) at the same point. Fig. 1.b illustrates the procedure for updating the values in the accumulator for the instantaneous case in which the three edge pixels of the image space shown in Figure 1a are analyzed.

Based on the above, this method involves the following steps:

(a) to obtain the image's edge map using an edge detector

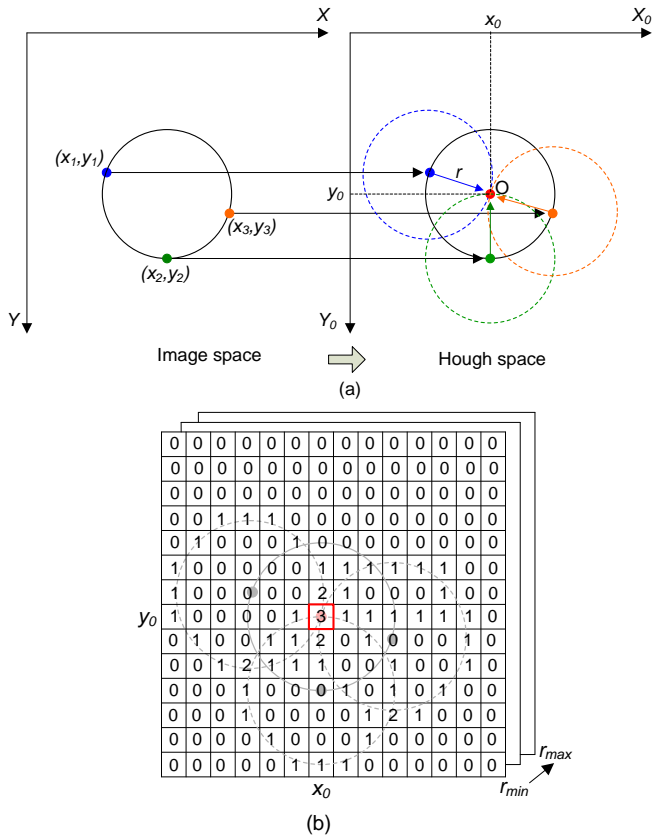

**Fig. 1.** (a) Principle of the Circular Hough Transform (b) State of the accumulator arrangement after the votes of the three points shown in (a).

(b) to explore each pixel in the image. If it is labeled as an edge, it will yield a circle of radius $r$ centered on itself. Cells belonging to the circle receive one vote.

(c) to determine the most voted cell which will correspond to the center of the circle of the image space.

CHT is used to detect the edge of the pupil and therefore the radius and its central coordinates in the image. To find the circle that best fits the contour of the pupil, the CHT algorithm is applied for different radius values in a range of radii from $r_{min}$ to $r_{max}$ that includes the estimated radius of the pupil. The pseudocode implemented for this method is shown below:

---
Pseudocode: CHT Algorithm
---
*Stage 1: Pre-processing*
1. Load image
2. Reduce image dimensions by a factor of 4.
*Stage 2: Detection using CHT*
3. Initialize to 0, the three-dimensional accumulator array of the Hough parameter space.
4. Detect the edges in the image using Canny's algorithm.
5. For each pixel in the image:
   For each radius from $r_{min}= 5$ to $r_{max}=25$
      If the point is on an edge and it meet that
      $g(x_j, y_j, x_0, y_0, r) = 0$
         Increase by one the elements of the accumulator array
6. Find the maximum in the accumulator array
7. Return the circle that corresponds to the maximum value found in the accumulator.

---

### 2.2 Ellipse Fitting

The Ellipse Fitting (EF) method is based on detecting the points located on the pupil contour and obtaining the ellipse that best fits these points, according to the least-squares criterion. Various algorithms implement this method depending on the variant used for contour detection.

In this work the algorithm proposed in (29) is implemented, which uses the algorithm presented in (30) for the ellipse fitting, since it offers a good tradeoff between speed and precision. The algorithm consists of two stages: pre-processing and fitting.

*Pre-processing*

Initially, the IR ocular image ($I_{input}$) is read, which represents the input of the algorithm, and a non-linear transformation is applied to obtain an output binary image ($I_{output}$). For it, the threshold value $T = 25$ was used so that:

$I_{output}(x,y) = 1$ if $I_{input}(x,y) > T$

$I_{output}(x,y) = 0$ if $I_{input}(x,y) \leq T$

The morphological closing operation (dilation followed by erosion) is then performed to reduce the noise effects caused by the eyelashes and other undesirable disturbances present in the binary image obtained in the previous step. This operation tends to smooth the contours of the objects, fuses narrow breaks and long thin gulfs, eliminates small holes, and fills gaps in the contour (31). The closing operation was implemented with a structural element in the form of a disk with radio 5.

Edge detection is then performed using Canny's algorithm (20) and, for the edge segments obtained from this method, an analysis of connected components is implemented in a neighborhood of 8 pixels. At this point, the objective is to detect connected components, which will be those neighboring regions or areas whose pixels are connected by a path or set of pixels of the same value (for example, 1) to which the same identification label will be assigned. The algorithm will remove the shortest connected component chains and store, in a two-dimensional matrix, the set of positions ($x$, $y$) of the pixels belonging to the longest connected component chain. The previously saved positions ($x$, $y$) will constitute the input parameters for the ellipse fitting function that is proposed as the second step of the method and is explained below.

*Fitting*

The fitting finds the parameters that define an ellipse in a sparse data set. The ellipse fitting algorithm (30) receives as input data a vector with the ($x$, $y$) coordinates resulting from the previous step. The data is then normalized to position the center of the ellipse at the coordinate origin.

Any conic, in general, can be represented by a second-order polynomial like the following:

$$F(a,x_i) = a \cdot x = ax^2 + bxy + cy^2 + dx + ey + f = 0 \quad (2)$$

where $a = [a\ b\ c\ d\ e\ f]^T$ and $x = [x^2\ xy\ y^2\ x\ y\ 1]^T$,

$F(a,x_i)$ is the so-called "algebraic distance" from one point ($x$, $y$) to the conic $F(a,x) = 0$.

The fitting of a general conic can be addressed by minimizing the sum of the square algebraic distances of the curve for the $N$ points.

$$D_A(a) = \sum_{i=1}^{N} F(x_i)^2 \quad (3)$$

To avoid the trivial solution $a = 0$, constraints $D$ are applied to vector $a$. The minimization of distances can be resolved considering the generalized eigenvalue system:

$$D^T D a = \lambda C a \quad (4)$$

Where:

$D = [x_1\ x_2\ ...\ x_n]^T$ is the design matrix

$D^T D$ is the dispersion matrix

$C$ is the 6x6 constraint matrix.

In the specific case of the ellipse, the constraint is quadratic in form $4ac - b^2 = 1$ and can be expressed in the matrix form $a^T C = 1$ as:

$$a^T \begin{pmatrix} 0 & 0 & 2 & 0 & 0 & 0 \\ 0 & -1 & 0 & 0 & 0 & 0 \\ 0 & 0 & 0 & 0 & 0 & 0 \\ 0 & 0 & 0 & 0 & 0 & 0 \\ 0 & 0 & 0 & 0 & 0 & 0 \\ 0 & 0 & 0 & 0 & 0 & 0 \end{pmatrix} a = 1 \quad (5)$$

With the above equations and constraints, we construct and solve the system, which has 6 pairs of eigenvalues and eigenvectors. For more details on the mathematical basis, see (30).

Finally, the fitting function returns the following parameters: center of the ellipse, radius, and orientation. With this data it is possible to trace the fitted ellipse to the pupil and show its center.

The pseudo-code implemented for this method is shown below:

| Pseudocode: EF Algorithm |
| --- |
| *Stage 1: Pre-processing* |
| 1. Read input image |
| 2. Convert to binary image |
| 3. Perform closing morphological operation |
| 4. Detect edges in the image |
| 5. Save positions ($x, y$) of the longest connected component chain |
| *Stage 2: Ellipse Fitting* |
| 6. Read vector of ($x, y$) |
| 7. Normalize the data |
| 8. Build design matrix |
| 9. Construct dispersion matrix. |
| 10. Construct 6×6 constraint matrix |
| 11. Solve the generalized eigenvalue system |
| 12. Obtain the fitting parameters |
| 13. Pupil detection |

## 2.3 Daugman's Integro-Differential Operator

In 2004, Daugman proposed the Integro-Differential Operator (IDO) based method for detecting the contour of the pupil and the iris (22), which assumes that the edges of the pupil and the iris are circular and searches, in an image smoothed by a Gaussian filter, the parameters (center and radius) of a circular edge on which the integral derivative is maximum.

Mathematically it is described by the expression:

$$\max_{(r,x_0,y_0)} \left| G_\sigma(r) * \frac{\partial}{\partial r} \oint_{r,x_0,y_0} \frac{I(x,y)}{2\pi r} ds \right| \qquad (6)$$

Where the symbol $*$ denotes convolution, $I(x,y)$ is the intensity of the pixel in the coordinates of the ocular image, $r$ is the radius of several circular regions, with centers in ($x_0, y_0$), on which the gradient is calculated. $G_\sigma$ is a Gaussian smoothing function with a spatial scale value and is mathematically described by expression 7:

$$G_\sigma(r) = \left(\frac{1}{2\pi\sigma}\right) e^{-\frac{(r-r_0)}{2\sigma^2}} \qquad (7)$$

The full operator behaves like a circular edge detector that iteratively searches for the edge that maximizes the IDO value within a circle with radius and center which is the integration surface. This operator is applied iteratively with the amount of smoothing progressively reduced to achieve the exact location.

Figure 2a illustrates the operating principle of the Daugman IDO, from the exploration of two pixels located in columns 6 and 14 (fifth row) of the image matrix $I$, which stores a binary image composed of a gray circle, radius $r_3$ equal to 3 pixels, on a white background. For convenience in the figure, it will be assumed that the pixels that have the gray color (partial or total) have an intensity equal to 1, and those that appear white, have an intensity equal to 0. In pixels with centers in ($x_{06}$, $y_{05}$) and ($x_{014}$, $y_{05}$), the line integral is calculated for each of the radii $r_1$, $r_2$, $r_3$ and $r_4$, equal to 1, 2, 3 and 4 pixels, respectively, and in each case it is divide (normalize) by the perimeter of each circle. Then the partial derivative concerning the radius is calculated from the results obtained previously and the modular value is obtained. The maximum operator value (1.06 in this case, figure 2b) corresponds to the radius of the circle sought, and the pixel coordinates correspond to the center of that circle: ($x_{06}$, $y_{05}$) in this case. The exploration of the pixel with coordinates ($x_{014}$, $y_{05}$), belonging to a point in the background of the image, does not show variations in the value of the integro-differential operator for any radius, so the existence of a circle is discarded in this case.

The upper and lower figure 2b shows the results obtained from the calculation of the integral and the derivative, respectively, for both example pixels and each radius. Notice that for each radius, the results are stored in the homologous row, of a three-dimensional matrix.

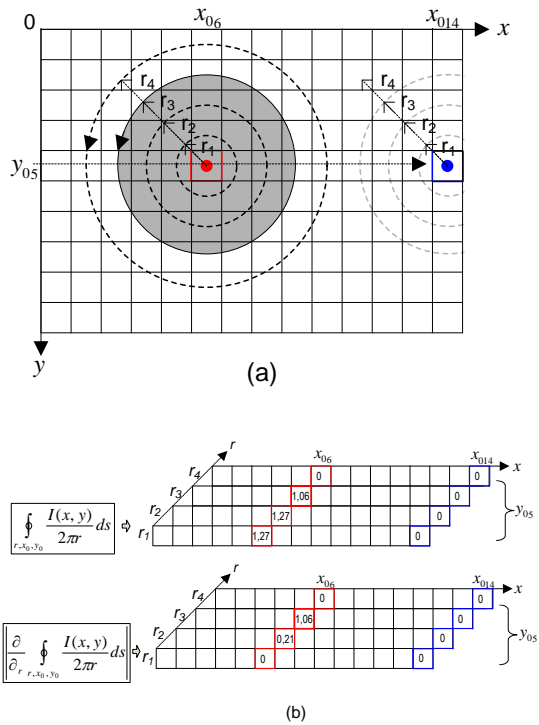

**Fig. 2.** Daugman's Integro-Differential Operator Principle. (a) Two pixels scan of columns 6 and 14 (fifth row) of image matrix *I* for radii equal to 1, 2, 3 and 4 pixels. (b) Results in row 5 of the integral (upper matrix).

In order to explain the method easily, in this case, the convolution operation with a Gaussian filter has not been taken into account, since it is used to eliminate the effects of noise in an image and it is being considered an ideal image without noise.

The implemented algorithm is the one proposed in (32) that optimizes the integro-differential operator by including a previous pre-processing stage whose function is to reduce the number of pixels of objects to which the Daugman's operator is applied. The algorithm, therefore, consists of two stages: pre-processing and detection of the pupil center using the IDO. In this case, unlike the original proposal, a range of radii corresponding to those of the pupil is used, and the iris is not detected.

*Pre-processing*

Initially, the input image is read and its dimensions are reduced, employing a subsampling, to decrease the computational cost. Since the Daugman's operator is very sensitive to light reflections in eye images, which affect pupil edge detection, a morphological operator is used that fills in the light-affected regions with the average light intensity pixels of the surrounding region. Then the grayscale image is converted to a binary image using a threshold of 25. This operation is applied after the previous step to mark as "object pixels", those that could be the central pixels (corresponding to the region of the pupil). Therefore, all pixels whose intensity is smaller than a threshold are marked, and the Daugman's operator applies only to those pixels. The image resulting from this transformation is then scanned pixel by pixel to determine if the pixel being analyzed represents a local minimum in its immediate neighborhood of 3×3. This means that the intensity of each pixel is compared with the intensities of its nine immediate neighboring pixels. The pixel with the lowest intensity value among these nine neighbors is used for other calculations and the rest of the pixels are discarded. Reducing the number of pixels in objects, in which the Daugman's operator is applied, diminishes the number of calculations and speeds up the detection process.

*Detection of the pupil center using the integro-differential operator*

To the sub-sampled image, from which pixels have been eliminated, the integro-differential operator is applied to detect the center and radius of the pupil, searching in a radius range of 5 to 25 pixels until a maximum is found, as explained above.

The pseudo-code implemented for this method is shown below:

| Pseudocode: IDO Algorithm |
|---|
| *Stage 1: Pre-processing* |
| 1. Read input image |
| 2. Reduce image dimensions by a factor of 4 |
| 3. Remove light spots |
| 4. Find local minimum in the neighborhood of a pixel |
| 5. Discard unrepresentative pixels |
| *Stage 2: Detection using IDO* |
| 6. Apply Gaussian filter |
| 7. Initialize pupil center and radius |
| 8. For each pixel in the image |
|    8.1 Construct circle with given center and radius |
|    8.2 Calculate integro-differential operator |
|    8.3 If the operator is maximum: |
|       - Set maximum operator value |
|     Else |
|       - Change center and radius |
| 9. Pupil detection |

## 2.4 Radial Symmetry Transform

This method is based on considering all the possible circles that a border pixel can be part of. Each point on the edge of a circle votes along a line of possible radii and these lines intersect at the center of the circle, resulting in a peak. The Radial Symmetry Transform (RST) was proposed in (25). Figure 3 shows the steps to obtain it.

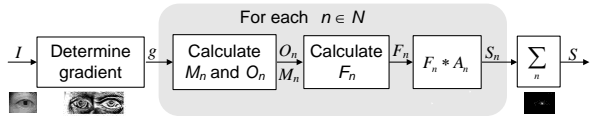

**Fig. 3.** Steps to obtain Radial Symmetry Transform.

The RST is calculated for one or more radii $n \in N$, where $N$ is the set of radii of the radially symmetrical characteristics to be detected. The value of the transform at radius $n$ indicates the contribution to the radial symmetry of the gradients at a distance $n$ from each point.

First, the image gradient $g$ is determined, which acts as an edge detector. If its value is positive, the radial symmetry contribution of each pixel with its surrounding pixels is analyzed. If not, the next pixel in the image is analyzed. Within neighbors, the value of the gradient of pairs of points symmetrically located above the central pixel is used as evidence of radial symmetry.

For each radius $n$, an orientation projection image $O_n$ and a magnitude projection image $M_n$ are calculated. These images are generated by examining the gradient $g$ in each pixel p from which a positively-affected pixel $p_{+ve}$ and a negatively-affected pixel $p_{-ve}$ are determined, as shown in Figure 4.

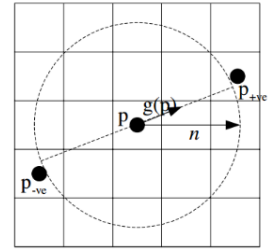

**Fig. 4.** Definition of positively-affected pixel $p_{+ve}$ and negatively-affected pixel $p_{-ve}$ by the gradient element $g(p)$ for a radius $n = 2$. The circle with dashed lines shows all the pixels that can be affected by the gradient for a radius $n$.

The positively-affected pixel is defined as the pixel where the gradient vector is pointing at a distance $n$ from $p$, and the negatively-affected pixel is defined as the pixel located at the same distance $n$ in the opposite direction to where the gradient is pointing.

The coordinates of the positively and negatively affected pixels are given by:

$$p_{+ve}(p) = p + round\left(\frac{g(p)}{\|g(p)\|}\right)n \qquad (8)$$

$$p_{-ve}(p) = p - round\left(\frac{g(p)}{\|g(p)\|}\right)n \qquad (9)$$

The "round" function rounds each element of the vector to the nearest integer. Orientation and magnitude projection images are initialized to zero. For each affected pair of pixels, the pixel corresponding to $p_{+ve}$ in the orientation projection image $O_n$ and in the magnitude projection image $M_n$ increases by 1 and $\|g(p)\|$, respectively, while the pixel corresponding to $p_{-ve}$ is reduced by these same amounts in each image. The above is expressed mathematically as:

$$O_n(p_{+ve}(p)) = O_n(p_{+ve}(p)) + 1 \qquad (10)$$

$$O_n(p_{-ve}(p)) = O_n(p_{-ve}(p)) - 1 \qquad (11)$$

$$M_n(p_{+ve}(p)) = M_n(p_{+ve}(p)) + \|g(p)\| \qquad (12)$$

$$M_n(p_{-ve}(p)) = M_n(p_{-ve}(p)) - \|g(p)\| \qquad (13)$$

The transform can be adjusted to find only dark or light regions of symmetry. To find symmetry exclusively in the dark regions (due to the characteristics of the pupil), when determining $M_n$ and $O_n$, only the negatively affected pixels should be considered. Thus, in the case of our application, the RST method is adjusted to implement only equations 11 and 13.

The radial symmetry contribution for radius $n$ is defined by the following convolution:

$$S_n = F_n * A_n \qquad (14)$$

$$F_n(p) = \frac{M_n(p)}{k_n} \left( \frac{|O_n(p)|}{k_n} \right)^\alpha \qquad (15)$$

Where $\alpha$ is the radial strictness parameter, $k_n$ is a scale factor that normalizes $M_n$ and $O_n$ to different radii, and $A_n$ is a two-dimensional Gaussian mask.

The radial strictness parameter $\alpha$ determines how strictly radial the symmetry must be for the RST to return a high value of interest, that is, it defines how radial the symmetry of the object must be. The value $\alpha=2$ was chosen, experimentally because it is a good tradeoff between rejection of non-radial symmetry elements (for example, eyelashes), accuracy, and computational cost. The normalization factor $k_n$ allows to compare (or combine) on the same scale, the symmetry images calculated for different radii.

To normalize, $O_n$ and $M_n$ are divided by their maximum values. The complete transformation $S$ is defined as the average of the contributions of symmetry over the entire set $N$ of radii considered, that is:

$$S = \frac{1}{|N|} \sum_{n \in N} S_n \qquad (16)$$

In consequence, to effectively detect the pupil center, the RST method leverages its symmetrical circular feature and calculates the negatively-affected pixel for each point on the image. Negatively-affected pixels point to the pupil center as they are located in the negative direction of the gradient, i.e. from a higher gray level (white) to a lower gray level (black). The final result of the RST is calculated according to the two projection images $M_n$ and $O_n$ which, according to the radius, are updated to save the contributions of the negatively-affected pixels. In the RST process, with the change of the radius values, the points of the pupil edge and of the iris edge will overlap near the center of the pupil, where a maximum value will result. The position of the RST maximum value is the location of the pupil center. The pseudo-code of the implemented algorithm is shown below:

| Pseudocode: RST Algorithm |
|---|

*Stage 1: Pre-processing*
1. Read input image
2. Sub-sample the image by a factor of 4

*Stage 2: Detection using RST*
3. Calculate image gradient
4. Initialize set of detection radii ($N_{min}= 5$, $N_{max}= 25$)
5. For each pixel in the image
6. For $n = N_{min}$:1: $N_{max}$
   - Calculate $p_{-ve}$ coordinates
   - Calculate $O_n(p_{-ve}(p))$ and $M_n(p_{-ve}(p))$
   - Calculate $F_n(p)$ according to equation 15
   - Calculate the RST result by equation 14 in the detection radius $n$, where the variance is chosen $\sigma = 0.1n$ and the size of the Gauss window is $ceil\left(\frac{n}{2}\right) \times ceil\left(\frac{n}{2}\right)^*$

7. Calculate $S$ by equation 16
8. Find in $S$ the coordinates of the maximum value $S_{max}$ that correspond to the location of the pupil center

## 3 Results and Discussion

To perform the comparative analysis, 800 IR images from the CASIA-IrisV3 and CASIA-IrisV4 databases (18) were used. They were randomly selected: 200 from the CASIA-IrisV3-Lamp, 200 from the CASIA-IrisV3-Twins, and 400 images from the CASIA-IrisV4-Thousand, all in JPG format and a resolution of 640 × 480 pixels. The CASIA-IrisV3-Lamp subset contains images affected by variations in lighting induced by a lamp, with consequent cases of pupil contraction and dilation. The CASIA-IrisV3-Twins subset contains images of twins, with noisy elements (interference of eyelashes, hair, and eyelids). The CASIA-IrisV3-Thousand subset contains good quality images of subjects with glasses and specular reflections.

The evaluation of the implemented methods was performed on a computer equipped with 3.1 GHz Intel Core I5 4670S microprocessor, 4 GB of DDR3 RAM, and Windows 7 Professional operating system for 64-bit architecture. Detection algorithms were applied to all images. The evaluation procedure covers the following steps:
- Analysis of accuracy
- Analysis of robustness
- Analysis of computational cost

**Accuracy analysis**

For the total set of images, the hit rate is defined as:

$$T_{hits} = \frac{P_c}{T_i} \cdot 100\% \qquad (17)$$

where: $P_c$ is the total of the pupils detected correctly and $T_i$ is the total of images.

For all the evaluated images, a specialist annotated the geometric center of the pupil using the PUPILA2.EXE tool (33) developed in our group. This tool contains several options for manual and semi-automatic annotation, which makes the labeling process more user friendly. The annotated coordinates constitute the reference value. A correct pupil detection is considered if the error ($e$) between the estimated center and the annotated center is less than or equal to 25% of the Euclidean distance ($d$) between these centers, divided by the radius of the pupil $R$, what which is expressed mathematically by the following expression:

$$e \leq \frac{0{,}25d}{R} \cdot 100\% \qquad (18)$$

The 25 % criterion was empirically determined, based on a previous analysis of the 800 experimental images. Since this metric is relational, it is more appropriate than considering that the Euclidean distance is less than 5 or 6 pixels (14,26). In addition, this metric considers the dilation and contraction of the pupil and it is more invariant to scale when the distance at which the eye image is captured is not the same.

Table 1 shows the results of the hit rate for each subset of the CASIA database after applying the algorithms described in the previous section. The method that showed the best performance, in terms of accuracy, was the Radial Symmetry Transform with a global hit rate for all images of 94.62%, followed by the methods of the Integro-Differential Operator, the Circular Hough Transform and the Ellipse Fitting with hit rates of 86.87%, 77% and 64.25%, respectively.

**Table 1.** Hit rate for the three subgroups of the CASIA database (EF: Ellipse Fitting, CHT: Circular Hough Transform, IDO: Integro-Differential Operator, Radial Symmetry Transform).

| Subset/ Algorithm | EF | CHT | IDO | RST |
|---|---|---|---|---|
| CASIA-IrisV4Thousand | 242/400 | 331/400 | 387/400 | 394/400 |
| CASIA-Iris V3 Twins | 146/200 | 143/200 | 128/200 | 175/200 |
| CASIA-Iris V3 Lamp | 129/200 | 142/200 | 180/200 | 188/200 |
| **Global hit rate (%)** | **64.25** | **77** | **86.87** | **94.62** |

**Robustness analysis**

The robustness of each method is quantified by analyzing the behavior of its hit rate in images with and without disturbances. For this, the 800 images of CASIA were subdivided into: 473 clear images, 136 images influenced by hair and eyelashes, 91 images influenced by the eyelid, and 100 images influenced by eyeglasses and reflections.

Table 2 shows the results of applying the methods to each of the experimental subsets and it is observed that the algorithm based on the RST presented the highest hit rate. The hit rate decreased considerably in the presence of images with interference of hair and eyelashes over the eye.

**Table 2.** Hit rate for the different experiments for robustness analysis using the database (EF: Ellipse Fitting, CHT: Circular Hough Transform, IDO: Integro-Differential Operator, Radial Symmetry Transform).

| Algorithm/ Experiment | EF (%) | CHT (%) | IDO (%) | RST (%) |
|---|---|---|---|---|
| Clear images | 76.53 | 87.31 | 91.54 | 97.46 |
| Images influenced by hair and eyelashes | 58.08 | 65.44 | 72.05 | 86.76 |
| Images influenced by the eyelid | 51.64 | 47.25 | 75.82 | 95.60 |
| Images influenced by glasses and reflections | 26 | 68 | 92 | 97 |
| **Average robustness (%)** | **53.06** | **67** | **82.85** | **94.45** |

Similarly, the RST also reached the highest robustness (94.45%), followed by the IDO, CHT, and EF methods with hit rates of 82.85%, 67%, and 53.06%, respectively. Figure 5 shows the result of the pupil location for image S2050L02.jpg of the subset of clear images.

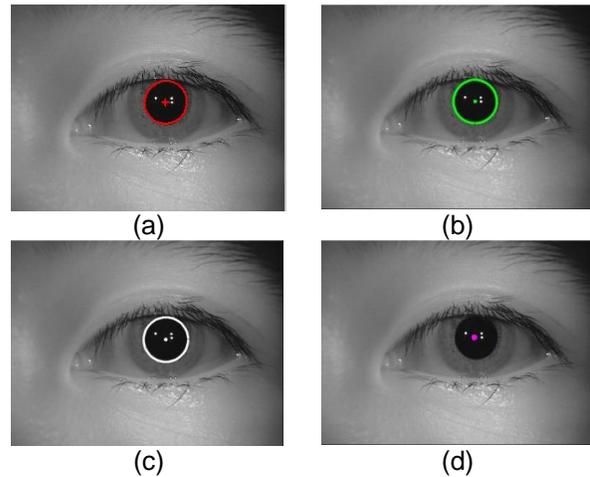

**Fig. 5.** Correct detection of the pupil center in the image S2050L02.jpg of the experiment with clear images: (a) Ellipse Fitting, (b) Circular Hough Transform, (c) Integro-differential Operator, (d) Radial Symmetry Transform.

Figure 6 shows the results of the algorithms for ocular images influenced by the presence of hair and eyelashes close to the region of interest (pupil). An incorrect detection was obtained when using the integro-differential operator.

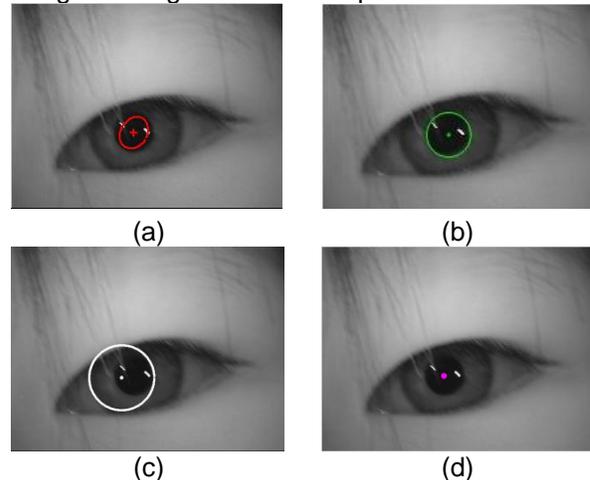

**Fig 6.** Detection of the pupil in the image S3191R01.jpg of the subset of ocular images influenced by hair and eyelashes. (a) Ellipse Fitting, (b) Circular Hough Transform, (c) Integro-Differential Operator, (d) Radial Symmetry Transform.

In the third experiment, corresponding to pupil detection in images influenced by the eyelid and the semi-occluded eye, the hit rate of the algorithms decreased considerably, especially for the CHT-based method. The detections made by the Ellipse-Fitting and Circular Hough Transform methods in Figure 7 (a) and (b) are considered erroneous.

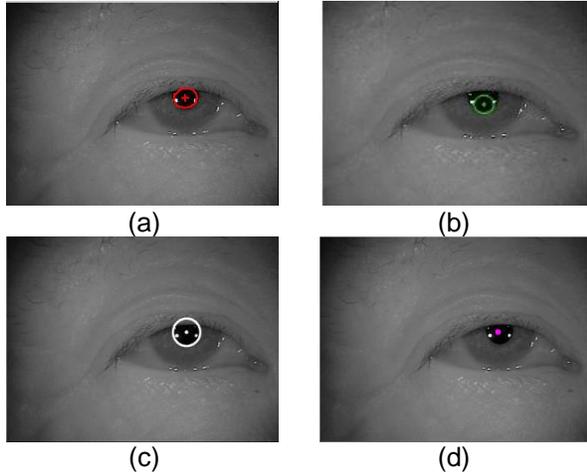

**Fig 7.** Detection of the pupil in the image S5559L00.jpg where the area of the pupil is partially covered. (a) Ellipse Fitting, (b) Circular Hough Transform, (c) Integro-Differential Operator, (d) Radial Symmetry Transform.

Finally, the subset of images influenced by glasses and reflections was processed. Figure 8 shows the results of the algorithms for image S5020L08.jpg from the CASIA-Iris-Thousand database.

### Computational cost analysis

The computational cost was estimated through the calculation of the execution time measured when processing the same image by each algorithm. For this, the MATLAB 2018 "tic-toc" function was used, which returns this time in seconds. This metric provides valuable information when it is desired to implement a VOG system in real time.

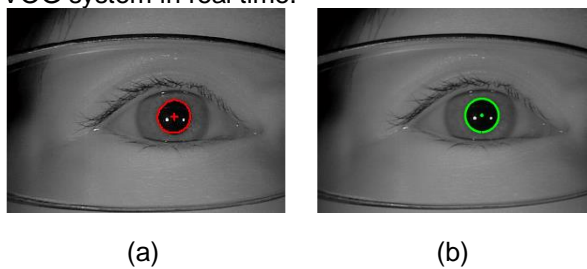

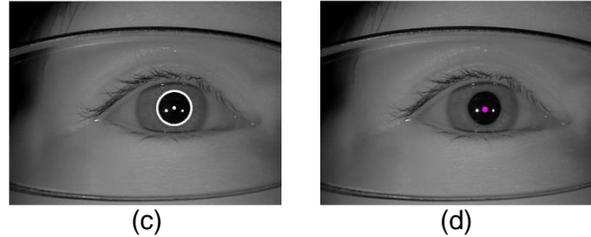

**Fig 8.** Pupil detection in the image S5020L08.jpg where the subject wears glasses, (a) Ellipse fitting, (b) Circular Hough Transform, (c) Integro-Differential Operator, (d) Radial Symmetry Transform.

Figure 9 shows the execution times for the four compared algorithms. We can appreciate that EF and CHT are the most efficient methods with less than 0.1 seconds each one. This information about the processing latency should be taken into account when applying a VOG system in practice.

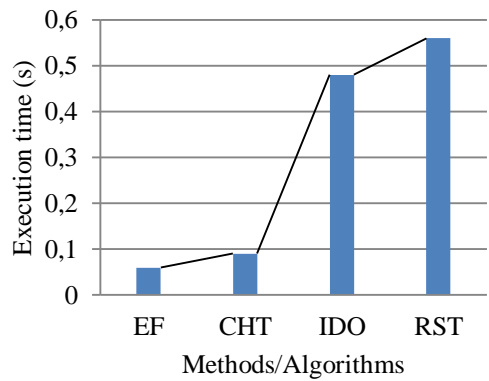

**Fig. 9.** Execution time (s) of the four algorithms

### General discussion

From the analysis of Table 2, it is observed that the level of robustness is maintained in most of the experiments in the following order (from highest to lowest): RST, IDO, CHT and EF, with the only exception that the EF was slightly higher than the CHT in images influenced by the eyelid, where the CHT showed its worst robustness. This last issue could be attributed to the fact that when the occlusion of the eye tends to increase, the EF, due to its foundation, requires fewer edge pixels (to estimate the pupil center within a certain error) than the CHT would require.

In general, the main problem with CHT is that the pupil is not a perfect circle and its shape and radius depend on different factors, such as the position of the pupil, lighting, corneal reflection, the presence of hair, and eyelashes, among others.

The EF showed very low robustness in the presence of glasses and reflections. This is because the use of thresholds causes the performance to vary depending on the characteristics of the image, sometimes being low. For example, in the case of eyeglasses with dark frame, thresholding may cause the longest connected component that the method seeks to be located on the frame and not on the pupil region. In general, the ellipse fitting method showed errors in images with low contrast and extreme variations in illumination.

Daugman's IDO presented the greatest difficulties in images influenced by hair, eyelashes and eyelid as these elements impair the circular contour of the pupil, affecting the value of the line integral in that contour and consequently the estimation of the center of the pupil. Failures were also registered when the contrast between the iris and the pupil is very low, which does not facilitate the correct segmentation of the pupil.

The RST algorithm proved to be robust in images where eyeglasses, high density of eyelashes, or flashes of light appear. Since the eyelashes and the frame of the glasses do not have radial symmetry, no matter how the detection radius changes, their results will not contribute to the detection of the center of the pupil. Similarly, since the light flares have a high gray level value (close to 255, white), the negative direction of the gradient at the points of the edge of the light flare deviates from the center of the circle, and will not represent a contribution to the results of the RST. Therefore, the light spot will also not affect the detection of the center of the pupil. Through experimental results, it was identified that the RST algorithm presents difficulty in processing images highly influenced by the eyelid, hair, and eyelashes. This result was expected considering that this occlusion in the pupil area destroys its circular character, which also affects the rest of the analyzed methods. In these cases, other alternative prediction methods should be investigated.

All methods achieve their best performance with clear images. Even though the RST and IDO had the best performance in accuracy and robustness, their processing times were relatively high compared to CHT and EF.

## 4 Conclusions

This work presented a quantitative and qualitative comparison on the performance of four of the most cited methods for pupil center estimation. Up to the best of our knowledge, no comparative works have been presented on this topic using equal conditions. Therefore, the novelty of this study lies in that it constitutes a first approximation in which the performance of the methods was assessed under equal conditions. This means using the same images, which are representative of real scenarios, the same computational resources, and the same performance metrics. The best performance in terms of accuracy and robustness was obtained by the method based on the radial symmetry transform, while the shortest processing time was achieved by the ellipse fitting method. This result suggest the use of one of these methods depending on the application. Although RST obtained the longest processing time, it can be implemented with reasonable resources taking into account current computing technologies. Future works should be focused on the implementation of this method in an efficient language as well as its optimization.